\documentclass[sigconf,screen]{acmart}
\usepackage{tikz}
\usepackage{graphicx}
\usetikzlibrary{shapes, arrows.meta, positioning, fit, backgrounds, decorations.pathmorphing, shadows, calc}
\AtBeginDocument{%
  }

\setcopyright{acmlicensed}
\copyrightyear{2018}
\acmYear{2018}
\acmDOI{XXXXXXX.XXXXXXX}
\acmConference[Conference acronym 'XX]{Make sure to enter the correct
  conference title from your rights confirmation email}{June 03--05,
  2018}{Woodstock, NY}
\acmISBN{978-1-4503-XXXX-X/2018/06}

\usepackage{tikz}
\usetikzlibrary{shapes, arrows.meta, positioning, fit, backgrounds, decorations.pathmorphing, calc}
\usepackage{graphicx} 
\usepackage{tikz}
\begin{document}

\title{CQSA: Byzantine-robust Clustered Quantum Secure Aggregation in Federated Learning}

\author{Arnab Nath}
\affiliation{%
  \institution{Indian Institute of Technology (BHU)}
  \city{Varanasi}
  \country{India}
}

\author{Harsh Kasyap}
\affiliation{%
  \institution{Indian Institute of Technology (BHU)}
  \city{Varanasi}
  \country{India}
}

\renewcommand{\shortauthors}{Arnab et al.}

\begin{abstract}
Federated Learning (FL) enables collaborative model training without sharing raw data. However, shared local model updates remain vulnerable to inference and poisoning attacks. Secure aggregation schemes have been proposed to mitigate these attacks. In this work, we aim to understand how these techniques are implemented in quantum-assisted FL. Quantum Secure Aggregation (QSA) has been proposed, offering information-theoretic privacy by encoding client updates into the global phase of multipartite entangled states. Existing QSA protocols, however, rely on a single global Greenberger–Horne–Zeilinger (GHZ) state shared among all participating clients. This design poses fundamental challenges: fidelity of large-scale GHZ states deteriorates rapidly with the increasing number of clients; and (ii) the global aggregation prevents the detection of Byzantine clients. We propose \textbf{C}lustered \textbf{Q}uantum \textbf{S}ecure \textbf{A}ggregation (CQSA), a modular aggregation framework that reconciles the physical constraints of near-term quantum hardware along with the need for Byzantine-robustness in FL. CQSA randomly partitions the clients into small clusters, each performing local quantum aggregation using high-fidelity, low-qubit GHZ states. The server analyzes statistical relationships between cluster-level aggregates employing common statistical measures such as cosine similarity and Euclidean distance to identify malicious contributions. Through theoretical analysis and simulations under depolarizing noise, we demonstrate that CQSA ensures stable model convergence, achieves superior state fidelity over global QSA.
\end{abstract}



\received{20 February 2007}
\received[revised]{12 March 2009}
\received[accepted]{5 June 2009}

\maketitle

\section{Introduction}
Federated Learning (FL) has established itself as a promising paradigm for collaborative privacy-preserving learning, enabling distributed clients to train a global model without sharing raw data~\cite{mcmahan2017communication}. By keeping data local and exchanging only model updates, FL mitigates direct privacy risks. However, recent studies have demonstrated that raw model updates can also leak sensitive information through reconstruction attacks such as model inversion and gradient leakage~\cite{mothukuri2021survey,nasr2019comprehensive,fu2022label,hu2021source}, where an adversary can recover pixel-perfect training data. To mitigate this, Secure Aggregation (SA) protocols~\cite{bonawitz2017practical,mansouri2023sok} have been developed to ensure that the central server receives only the algebraic sum of local updates, theoretically blinding it to individual contributions.

While classical SA approaches based on Multi-Party Computation (MPC) or Homomorphic Encryption (HE) provide strong privacy guarantees, they suffer from critical limitations in both efficiency and security integrity~\cite{bonawitz2017practical,jin2023fedml,ma2022privacy,du2023efficient}. First, they incur high communication overhead and computational latency, scaling poorly with the number of clients ($N$) and model dimension ($d$). Second, classical SA is inherently fragile against Byzantine attacks. This is because the protocol mathematically blinds the server to individual inputs; it simultaneously blinds the server to detect malicious activity. A ``poisoned'' update injected by a malicious client is aggregated indistinguishably from honest updates, corrupting the global model. Classical methods~\cite{roy2022eiffel,rathee2023elsa,bell2023acorn} to detect such attacks often require complex, heavy cryptographic proofs (e.g., Zero-Knowledge Proofs) that can be impractical for resource-constrained edge devices.

Quantum Secure Aggregation (QSA) has recently emerged as an alternative to address these bottlenecks~\cite{zhang2022federated}. Leveraging the principles of quantum mechanics, specifically entanglement and the no-cloning theorem~\cite{wootters1982nocloning}, QSA provides information-theoretic security of individual updates under ideal quantum channels, with communication complexity independent of model dimension. Protocols utilizing Greenberger--Horne--Zeilinger (GHZ) states allow for the encoding of classical parameters into quantum phases, enabling the server to measure the global aggregate directly without ever accessing individual values~\cite{zhang2022federated}.

Despite this promise, existing QSA frameworks face two critical issues that make their FL deployment difficult in the current Noisy Intermediate-Scale Quantum (NISQ) era:

--\textit{The Fidelity Constraint}. State-of-the-art QSA protocols typically assume a global entanglement scheme where all $N$ clients share a single $N$-qubit GHZ state. In practice, the fidelity of such large-scale entangled states decays exponentially with $N$ due to environmental decoherence and imperfect gate operations. Maintaining a stable entangled state across a large federated network (e.g., $N > 50$) is physically infeasible, resulting in noisy aggregates that degrade model convergence.

--\textit{The Verification Blind Spot.} 
After aggregation, the server learns only the final sum, but lacks the visibility to detect ``poisoned'' updates injected by malicious clients. If a subset of clients uploads updates designed to corrupt the global model, the server has no mathematical basis to identify or reject them, as the constituent parts of the sum are hidden by quantum superposition.

\noindent\textbf{Our work.} In this paper, we propose \textbf{C}lustered \textbf{Q}uantum \textbf{S}ecure \textbf{A}ggregation (CQSA), a novel aggregation scheme in FL, that bridges the gap between NISQ limitations and robust security requirements. Instead of relying on a fragile global entangled state, we partition the client network into small, manageable clusters (e.g., of sizes 4 or 5). This modular approach offers two key advantages: (1) it respects physical hardware constraints, generating high-fidelity GHZ states for small clusters, which is well within the coherence limits of current quantum processors; (2) By aggregating partial sums at the cluster level, the server can perform \textit{Inter-Cluster Verification}, employing existing Byzantine-robust aggregation schemes (utilising statistical operations such as Euclidean norm or cosine similarity) among cluster aggregated updates to identify and reject malicious groups before they impact the global model. 

The key contributions of this paper are as follows:
\begin{itemize}
  \item \textbf{NISQ-Compatible Architecture:} We introduce a clustered aggregation protocol that replaces large-$N$ GHZ states with small-$k$ GHZ states, significantly improving state fidelity and aggregation accuracy on near-term quantum devices.
  \item \textbf{Inter-Cluster Verification:} We propose a hybrid verification mechanism where the server can employ existing Byzantine-robust aggregation to compare cluster-level aggregates to detect poisoned updates, a capability absent in existing QSA.
  \item \textbf{Theoretical Analysis and Simulation:} We propose a hybrid verification mechanism that allows the server to employ Byzantine-robust aggregation to audit cluster-level aggregates. Further, we demonstrate through simulations that CQSA achieves superior state fidelity compared to global-aggregation protocols in NISQ environments.
\end{itemize}

\section{Preliminaries and Related Work}
\subsection{Federated Learning}
Federated Learning (FL) is a distributed machine learning paradigm, where multiple clients collaboratively train a shared global model, keeping the data on-premises~\cite{mcmahan2017communication}. Let $D_i$ denote the private dataset of client $i\in N$. In each communication round $t$, the server broadcasts the current global model parameters $w_t \in \mathbb{R}^d$ to a subset of clients $\mathcal{S}_t \subseteq \{1,\ldots,N\}$. Each participating client performs local model training and computes a model update $\Delta w_i^t$. The server updates the global model by aggregating the received updates as
\begin{equation}
w_{t+1} = w_t + \mathcal{A}\big( \{\Delta w_i^t\}_{i \in \mathcal{S}_t} \big),
\end{equation}
where $\mathcal{A}(\cdot)$ denotes an aggregation function, commonly a weighted average as in FedAvg~\cite{mcmahan2016federated}.

\subsubsection{Secure Aggregation}
Although FL avoids direct data sharing, the shared local model updates may still leak sensitive information. Secure Aggregation protocols are designed to protect client privacy by ensuring that the server learns only the aggregated update. Given local updates $\{\Delta w_i^t\}_{i \in \mathcal{S}_t}$, a secure aggregation protocol guarantees that the server obtains $\sum_{i \in \mathcal{S}_t} \Delta w_i^t$, while each individual update remains hidden.

Secure aggregation schemes are designed using Multi-Party Computation (MPC)~\cite{bonawitz2017practical,roy2022eiffel,rathee2023elsa}, Homomorphic Encryption (HE)~\cite{jin2023fedml,ma2022privacy,du2023efficient}, differential privacy (DP)~\cite{wei2020federated} and other encoding mechanisms. While these techniques effectively mitigate inference and reconstruction attacks, they also introduce significant computational and communication overhead. Moreover, secure aggregation can also blind the server to individual updates, thereby limiting the server's ability to detect Byzantine behavior.


\subsection{Quantum States and Qubits}
The fundamental unit of quantum information is the qubit. Unlike a classical bit that exists in a state of either 0 or 1, a qubit exists in a superposition of two orthogonal basis states, denoted as $\lvert 0 \rangle$ and $\lvert 1 \rangle$ (Dirac notation). A single-qubit state $\lvert \psi \rangle$ is defined as a linear combination of these basis vectors:
\[
\lvert \psi \rangle = \alpha \lvert 0 \rangle + \beta \lvert 1 \rangle,
\]
where $\alpha,\beta \in \mathbb{C}$ are probability amplitudes satisfying the normalization condition $\lvert \alpha \rvert^2 + \lvert \beta \rvert^2 = 1$. The state of the system collapses to $\lvert 0 \rangle$ or $\lvert 1 \rangle$ upon measurement with probabilities $\lvert \alpha \rvert^2$ and $\lvert \beta \rvert^2$, respectively.

For a system of $N$ qubits, the state space grows exponentially, represented by the tensor product of individual Hilbert spaces. A general $N$-qubit state can be written as a superposition of $2^N$ computational basis states.

\subsection{Greenberger--Horne--Zeilinger State}
Greenberger--Horne--Zeilinger (GHZ) state is a maximally entangled quantum state that exhibits strong non-local correlations. For a system of $N$ parties, the $N$-qubit GHZ state is defined as an equal superposition of the ``all-zero'' and ``all-one'' states:
\[
\lvert \mathrm{GHZ} \rangle_{N} = \frac{1}{\sqrt{2}}\left( \lvert 0 \rangle^{\otimes N} + \lvert 1 \rangle^{\otimes N} \right),
\]
where $\lvert 0 \rangle^{\otimes N} \equiv \lvert 00\dots 0 \rangle$ and $\lvert 1 \rangle^{\otimes N} \equiv \lvert 11\dots 1 \rangle$. The crucial property of the GHZ state for secure aggregation is its sensitivity to local phase operations. An operation applied to any single qubit in the entangled system affects the global phase of the entire state, allowing for the aggregation of distributed information without collapsing the state into individual components.

\subsubsection{GHZ-Based Architecture: Mathematical States and Protocol}
We utilize a Quantum Secure Aggregation (QSA) protocol, which relies on the properties of GHZ states to perform blind summation. The protocol evolves the quantum system through three stages: initialization, encoding, and decoding.

--\textit{Initialization and Distribution.} The protocol begins with the aggregation server generating an $N$-qubit GHZ state, i.e., an equal superposition of the ``all-zero'' and ``all-one'' computational basis states:
\[
\lvert \Phi \rangle 
= \frac{1}{\sqrt{2}}\Bigl(\lvert 00\dots 0 \rangle + \lvert 11\dots 1 \rangle\Bigr).
\]
The server distributes these qubits such that each of the $N$ participants holds exactly one qubit of the entangled system.

--\textit{Private Encoding.} To submit a local update, each participant applies a phase gate (specifically an $R_z$ rotation) to their individual qubit using their private normalized parameter $\theta_i$. Because the qubits are entangled, these local rotations accumulate in the global relative phase of the system. After all $N$ participants have encoded their values, the global state becomes
\[
\lvert \Psi \rangle
= \frac{1}{\sqrt{2}}\left(\lvert 00\dots 0 \rangle + e^{i\sum_{n=1}^{N}\theta_n}\lvert 11\dots 1 \rangle\right).
\]
Then, the aggregate $\sum_{n=1}^{N}\theta_n$ is locked into the relative phase between the two basis states, while the individual values $\theta_i$ remain hidden.

--\textit{Decoding and Measurement.} The participants return their qubits to the server. To extract the sum, the server applies a decoding circuit consisting of a cascade of CNOT gates followed by a Hadamard gate on the first qubit, which disentangles the system and transfers the phase information into measurable probability amplitudes.

When the server measures the first qubit, the probabilities of observing outcomes $0$ and $1$ are determined by the aggregated phase $\Sigma = \sum_{n=1}^{N}\theta_n$:
\[
P(q_1=0) = \frac{1}{2}\bigl(1 + \cos(\Sigma)\bigr),
\qquad
P(q_1=1) = \frac{1}{2}\bigl(1 - \cos(\Sigma)\bigr).
\]
By repeating this measurement multiple times and calculating the frequency of ``0''s versus ``1''s, the server can invert these probability formulas to estimate the sum of the model updates with high precision.

\subsection{Related Work}

\subsubsection{Byzantine-Robust Aggregation}
Byzantine-robust aggregation techniques have been proposed to defend against malicious clients that submit poisoned updates. State-of-the-art approaches include \emph{Krum} and \emph{Multi-Krum}~\cite{blanchard2017machine}, which select updates that are closest to the majority by minimizing pairwise distances and are provably robust under bounded Byzantine assumptions. Coordinate-wise methods, such as the \emph{Median} and \emph{Trimmed Mean}~\cite{yin2018byzantine}, operate independently on each model parameter, thereby limiting the influence of extreme values. \emph{FLTrust}~\cite{cao2020fltrust} is another defense in line, which uses cosine similarity \textit{w.r.t.} a trusted root update to detect malicious updates. More recent approaches, such as \emph{MESAS}~\cite{krauss2023mesas}, combine multiple statistical indicators, including Euclidean distance, Cosine similarity, Variance, Min, Max, and Count. While these methods improve robustness against poisoning attacks, they rely on the server’s access to individual client updates in plaintext, making them incompatible with secure or privacy-preserving aggregation mechanisms.

\subsubsection{Quantum Secure Aggregation (QSA)} To address classical bottlenecks, Quantum Secure Aggregation has been introduced~\cite{zhang2022federated}. Protocols utilizing GHZ states allow the server to measure the global aggregate of model parameters directly via entanglement, achieving information-theoretic security~\cite{zhang2022federated}. Unlike classical HE, QSA can be computationally efficient and can detect eavesdropping on the quantum channel due to measurement disturbance~\cite{wootters1982nocloning}. However, existing QSA frameworks rely on a global GHZ state involving all participants. This architecture faces challenges regarding NISQ fidelity (due to decoherence in large states) and lacks intrinsic mechanisms to verify the integrity of the aggregated data against active Byzantine faults within the logical layer.

\section{Threat Model}
We consider an FL system under a comprehensive threat model that addresses privacy leakage, data integrity, and system reliability. We define the security assumptions for the three key entities: the central server, the participating clients, and external adversaries.

--\textit{Semi-honest Server.} We assume the central aggregation server $(\mathcal{S})$ is \textit{honest-but-curious}. The server strictly follows the established quantum aggregation protocol (e.g., preparing GHZ states and performing the prescribed measurements) but attempts to infer sensitive information about individual client gradients $(\mathbf{w}_i)$ from the aggregated outcomes or quantum residues. 

--\textit{Malicious (Byzantine) Clients.} We assume a subset of clients $\mathcal{B} \subset \mathcal{U}$ may be active adversaries. These malicious clients may deviate from the protocol by injecting random noise or specifically crafted poisoned gradients to prevent global model convergence.

--\textit{External Adversaries.} We assume external entities may intercept communication channels. These adversaries may engage in passive eavesdropping on quantum states or in active Man-in-the-Middle (MitM) attacks. The protocol relies on the no-cloning theorem and entanglement decoherence to detect such intrusions.




\begin{figure*}[!ht] 
\centering
\resizebox{0.9\textwidth}{!}{%
\begin{tikzpicture}[
    server/.style={
        rectangle, rounded corners, draw=blue!80!black, thick, 
        fill=blue!5, minimum width=14cm, minimum height=2.6cm, 
        align=center, drop shadow
    },
    module/.style={
        rectangle, draw=black!70, dashed, thick, 
        fill=white, minimum width=3cm, minimum height=1cm, 
        align=center, drop shadow, font=\footnotesize
    },
    defense/.style={
        rectangle, draw=red!80!black, thick, 
        fill=red!5, minimum width=3.5cm, minimum height=1.5cm, 
        align=center, drop shadow, font=\footnotesize\bfseries
    },
    client/.style={
        circle, draw=green!60!black, thick, fill=green!10, 
        minimum size=1.2cm, align=center, font=\footnotesize
    },
    malicious/.style={
        circle, draw=red!80!black, thick, fill=red!20, 
        minimum size=1.4cm, align=center, font=\footnotesize\bfseries, text=red!80!black
    },
    clusterbox/.style={
        draw=gray!50, dashed, thick, rounded corners=15pt, 
        inner sep=0.4cm, fill=gray!5
    },
    malicious_clusterbox/.style={
        draw=red!50, dashed, thick, rounded corners=15pt, 
        inner sep=0.4cm, fill=red!5
    },
    quantum_link/.style={
        -Latex, very thick, color=purple!80, 
        decorate, decoration={snake, amplitude=.4mm, segment length=2mm, post length=3mm}
    },
    classical_link/.style={
        Latex-Latex, thick, color=blue!70!black, dashed
    }
]

    \node[server] (FS) at (0,0) {};
    \node[below=0.1cm, font=\large\bfseries] at (FS.north) {CQSA Federated Server};

    \node[module] (agg) at ([xshift=-4cm, yshift=0cm]FS.center) {
        \textbf{Global Aggregation}\\
        $\Theta = \sum \Theta_j$
    };

    \node[defense] (filter) at ([xshift=3cm, yshift=0 cm]FS.center) {
        \textbf{Byzantine-robust}\\
        Aggregation\\
        (malicious weights not aggregated)
    };

    \draw[-Latex, thick] (filter) -- node[above, font=\scriptsize]{Verified} (agg);

    \node[below=3cm of FS] (mid) {};
    \node[client] (c2_1) at ([xshift=-1.2cm]mid) {$C_{2,1}$};
    \node[client] (c2_k) at ([xshift=1.2cm]mid) {$C_{2,k}$};
    \node at (mid) {$\dots$};
    \begin{scope}[on background layer]
        \node[clusterbox, fit=(c2_1) (c2_k), label={[gray, yshift=-0.8cm]below:\textbf{Cluster 2} (Honest)}] (G2) {};
    \end{scope}

    \node[client, left=4cm of c2_1] (c1_k) {$C_{1,k}$};
    \node[client, left=0.5cm of c1_k] (c1_dots) {$\dots$};
    \node[client, left=0.5cm of c1_dots] (c1_1) {$C_{1,1}$};
    \begin{scope}[on background layer]
        \node[clusterbox, fit=(c1_1) (c1_k), label={[gray, yshift=-0.8cm]below:\textbf{Cluster 1} (Honest)}] (G1) {};
    \end{scope}

    \node[client, right=4cm of c2_k] (cm_1) {$C_{M,1}$};
    \node[right=0.5cm of cm_1] (cm_dots) {$\dots$};
    \node[malicious, right=0.5cm of cm_dots] (cm_mal) {$C_{M,k}$\\Adv.};
    \begin{scope}[on background layer]
        \node[malicious_clusterbox, fit=(cm_1) (cm_mal), label={[red, yshift=-0.8cm]below:\textbf{Cluster M} (Compromised)}] (GM) {};
    \end{scope}


    \draw[quantum_link] (G1.north) -- ([xshift=-5cm, yshift=1cm ]FS.south) 
        node[midway, left=0.1cm, font=\scriptsize, align=right] {$\Theta_1$ (Clean)\\ \textit{Secure Upload}};
    
    \draw[quantum_link] (G2.north) -- (FS.south) 
        node[midway, fill=white, font=\scriptsize, inner sep=1pt] {$\Theta_2$};

    \draw[quantum_link, color=red!70] (GM.north) -- (filter.south) 
        node[midway, right=0.5cm, font=\scriptsize, align=left, color=red] {$\tilde{\Theta}_M$ (Poisoned)\\ \textit{Attack}};

    
    \draw[classical_link] (FS.west) to[out=180, in=110] node[midway, left, font=\scriptsize, align=right, color=blue!70!black] {Max Weight Broadcast\\ \& Coordination} (G1.north west);
    
    \draw[classical_link] (FS.east) to[out=0, in=70] node[midway, right, font=\scriptsize, align=left, color=blue!70!black] {} (GM.north east);

    \draw[classical_link] ([xshift=1cm]FS.south) -- ([xshift=1cm]G2.north);

    \node[draw=black!50, fill=white, rounded corners, anchor=south west] at ([xshift=0cm, yshift=0.5cm]c1_1.north west) {
        \scriptsize
        \begin{tabular}{l}
            \textbf{Legend:}\\
            \textcolor{purple}{\rule[2pt]{10pt}{2pt}} Quantum Channel (Secure Aggregation)\\
            \textcolor{blue!70!black}{- - -} Classical Channel (General Comm.)\\
            \textcolor{green!60!black}{$\bullet$} Honest Client \quad \textcolor{red!80!black}{$\bullet$} Malicious Client
        \end{tabular}
    };

\end{tikzpicture}
} 

\caption{\textbf{Comprehensive Architecture of the CQSA-FL System.} 
}
\label{fig:cqsa_architecture}
\end{figure*}
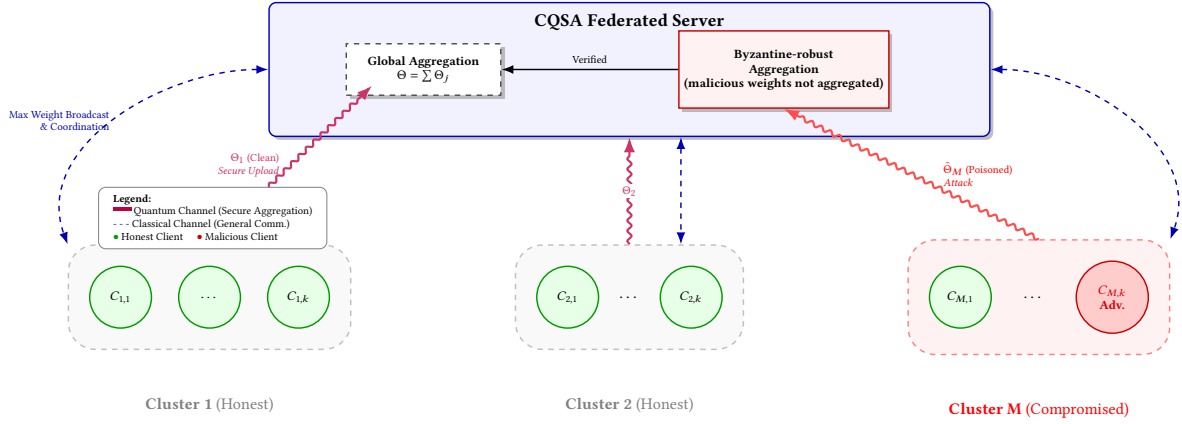

\section{CQSA: Our Proposed Aggregation Scheme}

\textbf{C}lustered \textbf{Q}uantum \textbf{S}ecure \textbf{A}ggregation (CQSA) is an advanced architectural approach to privacy-preserving Federated Learning (FL) that bridges the gap between theoretical quantum security and the practical constraints of the Noisy Intermediate-Scale Quantum (NISQ) era. Unlike traditional global Quantum Secure Aggregation (QSA), which attempts to entangle all $N$ participating clients into a single, fragile high-order GHZ state, CQSA partitions the network into smaller, manageable sub-clusters. Now, we discuss each component of CQSA in detail below.

\noindent\textbf{Encoding the weights into the phase.} 
Let $w_{i,j}$ denote the $i$-th coordinate of the $d$-dimensional model update of client $j$, which is mapped into the quantum phase domain. To ensure that aggregated phase values remain within the valid range $[-\pi, \pi]$, we introduce a scaling factor $S$ based on the maximum magnitude of the updates. Each client $j$ locally computes
$$w_{\max}^{(j)} = \max_{1 \le i \le d} |w_{i,j}|.$$
and transmits this scalar value to the server over a classical channel. The server then computes the global bound
$w_{\max} = \max\{w_{\max}^{(j)}\},$
which is broadcast back to all clients. This procedure reveals only a global magnitude bound and does not expose individual model updates or feature-level information.
The scaling factor is defined as:$$S = \frac{\pi}{k \cdot w_{\text{max}}}.$$The encoded quantum parameter (phase angle) $\theta_{i,j}$ for the $j$-th client is:$$\theta_{i,j} = S \cdot w_{i,j} = \frac{\pi \cdot w_{i,j}}{k \cdot w_{\text{max}}}.$$Note: The inclusion of $k$ in the denominator of $\theta_{i,j}$ is a necessary condition to prevent integer overflow (phase wrapping modulo $2\pi$) during the aggregation of $k$ clients. This linear mapping preserves the geometric properties (such as Euclidean norm and Cosine similarity) of the original weight space.

\noindent\textbf{Geometric Preservation in Quantum Phase Encoding.} We demonstrate that this linear mapping preserves the fundamental geometric properties of the weight space. Let $\boldsymbol{\theta_j}$ be the mapped weight vector of client \textit{j}.

--\textit{Proportionality of Euclidean Norm.} The Euclidean norm of the encoded phase vector is strictly proportional to the norm of the classical weight vector.

\begin{equation}
\begin{aligned}
    ||\boldsymbol{\theta}_j||_2 &= \sqrt{\sum_{j} (S \cdot \boldsymbol{w_j)}^2} 
    &= |S| \sqrt{\sum_{j} \boldsymbol{w_j}^2}.
\end{aligned}
\end{equation}

\begin{equation}
    \therefore ||\boldsymbol{\theta}_j||_2 = S \cdot ||\mathbf{w}_j||_2.
\end{equation}

\textit{Thus, the magnitude of the parameter vector in the quantum domain is directly proportional to the classical vector.}

--\textit{Invariance of Cosine Similarity}
The direction of the vector, measured by Cosine Similarity, is invariant under the scalar multiplication $S$. Let $\mathbf{v}$ be any reference vector (e.g., global model) and $\boldsymbol{\phi} = S \cdot \mathbf{v}$ be its encoded form.

\begin{equation}
    \text{CosSim}(\boldsymbol{\theta}_j, \boldsymbol{\phi}) = \frac{\boldsymbol{\theta}_j \cdot \boldsymbol{\phi}}{||\boldsymbol{\theta}_j||_2 \cdot ||\boldsymbol{\phi}||_2}.
\end{equation}

Substituting the mapping:

\begin{equation}
\begin{aligned}
    &= \frac{(S \mathbf{w}_j) \cdot (S \mathbf{v})}{(S ||\mathbf{w}_j||_2) (S ||\mathbf{v}||_2)}, \\
    &= \frac{S^2 (\mathbf{w}_j \cdot \mathbf{v})}{S^2 (||\mathbf{w}_j||_2 \cdot ||\mathbf{v}||_2)}.
\end{aligned}
\end{equation}

\begin{equation}
    \therefore \text{CosSim}(\boldsymbol{\theta}_j, \boldsymbol{\phi}) = \text{CosSim}(\mathbf{w}_j, \mathbf{v}).
\end{equation}

\textit{Theoretically, the directional alignment of the gradients is preserved in the quantum domain.}

\noindent\textbf{Client Partitioning and Randomization.} To circumvent the fidelity constraints of global entanglement, we partition the set of $N$ clients, denoted as $\mathcal{U} = \{C_1, C_2, \dots, C_N\}$, into $M$ disjoint clusters $\{G_1, G_2, \dots, G_M\}$. Each cluster $G_j$ consists of a small subset of $k$ clients. The partitioning is dynamic and randomized at the start of each training round to prevent adversaries from predicting cluster composition and colluding. We employ the Fisher--Yates shuffle algorithm to generate a random permutation of client indices, which are then sequentially sliced into groups of size $k$.


\noindent\textbf{Intra-Cluster Aggregation.} Within each cluster $G_j$, the clients perform the standard GHZ-based QSA protocol. Let $\boldsymbol{w_i}$ denote the local gradient weight of client $C_i$. This vector is normalized and encoded into a quantum phase angle $\boldsymbol{\theta_i} \in [-\pi,\pi]$. The aggregation server prepares a $k$-qubit GHZ state and distributes it to the members of $G_j$.

After encoding, the server decodes the state to obtain the cluster aggregate, denoted as $\mathbf{\Theta_j}$. Mathematically, this corresponds to the superposition of individual phases:
\[
\boldsymbol{\Theta_j} = \sum_{C_i \in G_j} \boldsymbol{\theta_i}.
\]

\textit{Privacy Guarantee.} It is important to note that the server only measures the aggregated value $\boldsymbol{\Theta_j}$. The individual contributions $\boldsymbol{\theta_i}$ remain locked in the entangled state and are destroyed upon measurement. Thus, the privacy of individual clients is preserved against the server.

\noindent\textbf{Inter-Cluster Verification.} Since the server possesses the set of cluster aggregates $\boldsymbol{\{\Theta_1, \Theta_2, \dots, \Theta_M\}}$, it can perform Inter-Cluster Verification to detect malicious (Byzantine) participation. 

From existing state-of-the-art Byzantine-robust aggregation schemes, it can be inferred that most schemes utilise statistical operations, such as Euclidean distance and cosine similarity, to detect outliers. Thus, we show that these metrics can also be calculated across aggregated clusters.

\begin{enumerate}
  \item \textbf{Cosine Similarity} ($S_{\mathrm{cos}}$): measures the directional alignment between two cluster aggregates $\boldsymbol{\Theta}_a$ and $\boldsymbol{\Theta}_b$,
  \[
  S_{\mathrm{cos}}(\boldsymbol{\Theta}_a, \boldsymbol{\Theta}_b) = \frac{\boldsymbol{\Theta}_a \cdot \boldsymbol{\Theta}_b}{\lVert \boldsymbol{\Theta}_a \rVert\, \lVert \boldsymbol{\Theta}_b \rVert}.
  \]
  
  \item \textbf{Euclidean Distance} ($D_{\mathrm{Euc}}$): measures the magnitude of difference between cluster aggregates,
  \[
  D_{\mathrm{Euc}}(\boldsymbol{\Theta}_a, \boldsymbol{\Theta}_b) = \sqrt{(\boldsymbol{\Theta}_a - \boldsymbol{\Theta}_b)^2} = \lvert \boldsymbol{\Theta}_a - \boldsymbol{\Theta}_b \rvert.
  \]
\end{enumerate}

\noindent\textbf{Outlier Rejection Logic.} The CQSA framework can be tailored with existing state-of-the-art Byzantine-robust aggregation schemes (e.g., Krum, Multi-Krum, Median, Trimmed Mean, FLTrust, MESAS, and more). 
Distance-based methods such as Krum and Multi-Krum can be instantiated by computing pairwise Euclidean distances between cluster-level updates and selecting those that are closest to the majority. Similarly, coordinate-wise robust statistics, including Median and Trimmed Mean, can be applied independently to each model parameter across clusters, limiting the influence of anomalous aggregates while preserving convergence behavior. CQSA also supports trust-aware and multi-statistic defenses such as FLTrust and MESAS. In this setting, the server evaluates cluster-level updates using similarity metrics (e.g., cosine similarity with a trusted reference update in FLTrust) and combined statistical indicators (e.g., Euc, Cos, Count, Variance, Min and Max measures in MESAS).

\noindent\textbf{Architectural Benefits.} The proposed CQSA architecture offers distinct physical and computational advantages over the standard global QSA approach.

--\textit{Increased State Fidelity.} Quantum state fidelity decays exponentially with the number of qubits due to environmental noise. For a global GHZ state of size $N$, the fidelity $F_N$ is approximately $F_N \approx (1 - \epsilon)^N$, where $\epsilon$ is the error rate per qubit. By reducing the entanglement size to $k$, we achieve a significantly higher fidelity:
\[
F_{\mathrm{cluster}} \approx (1 - \epsilon)^k \gg (1 - \epsilon)^N.
\]

--\textit{Reduced Coherence Time Requirements.}
The coherence time $T_2$ is the window during which a quantum state remains viable. Global aggregation requires all $N$ clients to encode and return their qubits within this window ($T_{\mathrm{op}} < T_2$). As $N$ grows, the transmission and wait times (often $T_{\mathrm{op}} \propto N$) inevitably exceed $T_2$, destroying the state. In our clustered approach, $T_{\mathrm{op}} \propto k$. Since $k$ is small and constant, the operation fits comfortably within the coherence limits ($T_{\mathrm{op}} \ll T_2$).

--\textit{Parallelization and Latency Reduction.} Global QSA is sequential in its preparation and measurement complexity. CQSA allows for parallelization. The $M$ clusters can be processed simultaneously by the server (assuming sufficient quantum hardware ports) or in rapid pipelined batches. The latency for a global round is reduced from $\mathcal{O}(N)$ to $\mathcal{O}(k)$.

--\textit{Localized Dropout Resilience.} In a global GHZ scheme, if a single client $C_i$ drops out (due to connectivity loss), the entire $N$-qubit entanglement is broken and the whole round must be restarted. In CQSA, a dropout only invalidates the specific cluster $G_j$. The server proceeds with the aggregation of the remaining $M-1$ clusters. 

\section{Evaluation}

\begin{figure}
    \centering
    \includegraphics[width=0.4\textwidth]{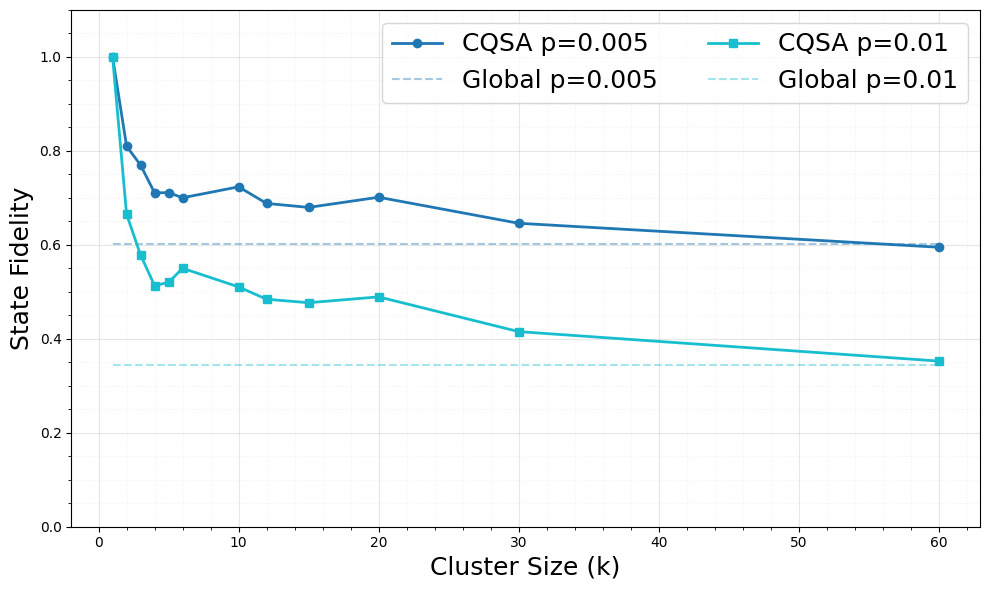}
    \caption{State Fidelity vs. Cluster size. The knobs in the solid line denote cluster size (factors of 60), starting from 1.}
    \label{fig:statefidelity}
\end{figure}

\begin{figure}
    \centering
    \includegraphics[width=0.4\textwidth]{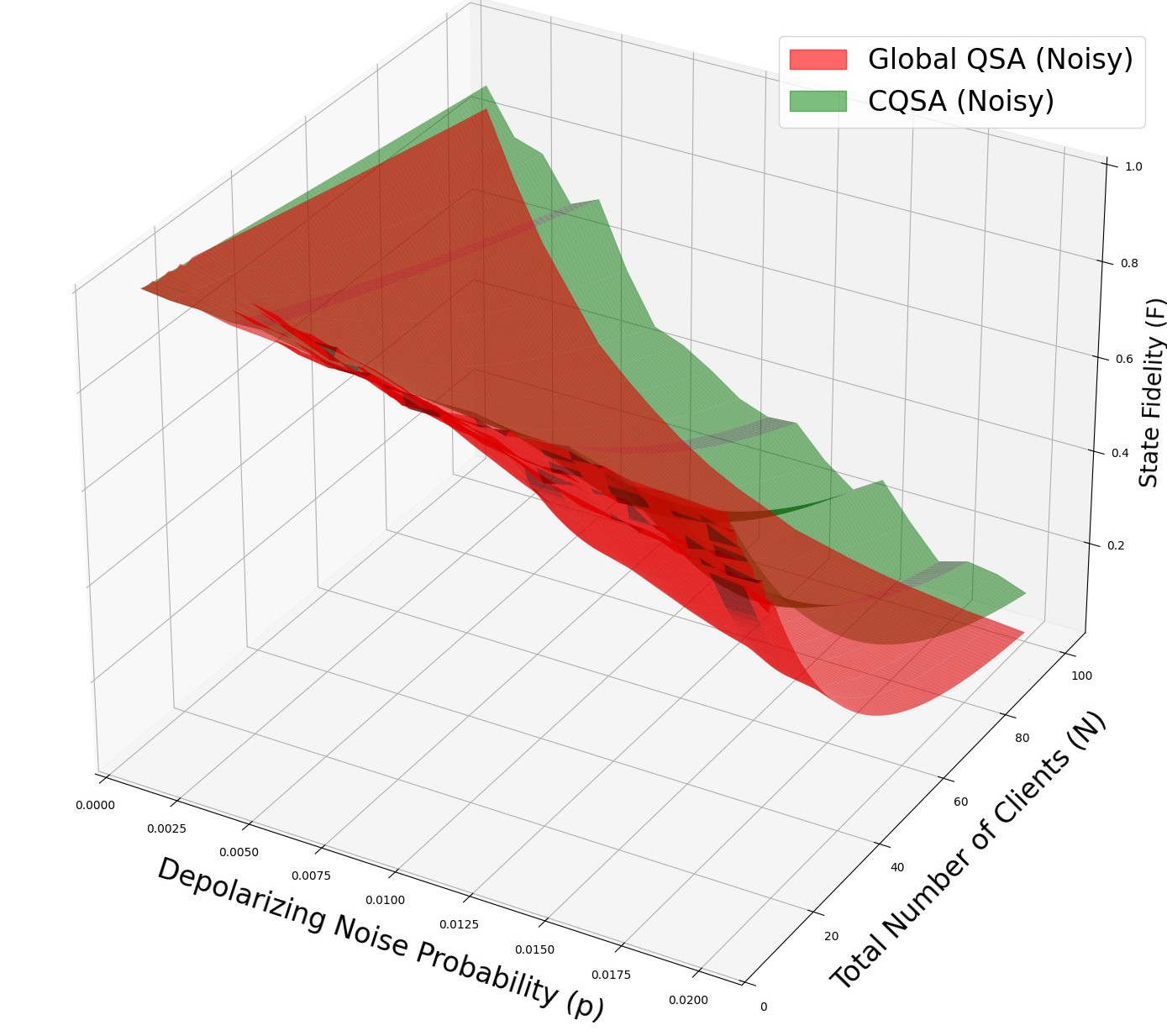}
    \caption{Comparative Fidelity Analysis across Noise and Scaling Dimensions (here cluster size k=4).}
    \label{fig:compfidelity}
\end{figure}

\noindent\textbf{Simulation Setup.} 
\textit{GHZ State Fidelity under Depolarizing Noise.} We evaluated the fidelity of distributed GHZ state generation using the NVIDIA CUDA-Q platform~\cite{nvidia2024cudaq}. To simulate realistic hardware imperfections, we utilized a local depolarizing noise model. A two-qubit depolarizing channel with probability ($p= 0.5\%$ for Figure~\ref{fig:statefidelity}, and range of 0 to 2\% for Figure~\ref{fig:compfidelity}) was applied exclusively to all CNOT gates in the circuit. Fidelity was defined as the population overlaps with the ideal computational basis states ($|0\dots0\rangle$ and $|1\dots1\rangle$). We employed a hybrid simulation technique to estimate scaling for large system sizes ($N=100$). For system sizes $n \leq 20$, fidelity was calculated via direct noisy simulation. For $n > 20$, fidelity was extrapolated based on the gate error rate using the recurrence relation $F_{n} = F_{n-1} \cdot (1-p)^2$. The CQSA metric models the system as $M = N/k$ independent clusters of size $k$. The total system fidelity is calculated as the product of the individual cluster fidelities, $F_{\text{total}} = (F_{\text{cluster}}(k))^{N/k}$, representing the joint probability that all distributed clusters simultaneously achieve the target state. \textit{We do not simulate FL with Byzantine-robust aggregation schemes explicitly, due to limited space. However, theoretical analysis proved that the statistical relationships are preserved; hence, CQSA can be tailored to existing schemes.}
\smallskip

\noindent\textbf{Simulation Result.} Figure~\ref{fig:statefidelity} demonstrates a clear inverse relationship between cluster size ($k$) and state fidelity: as the cluster size increases, the overall system fidelity significantly decreases. The Clustered QSA (CQSA) approach (solid lines) consistently outperforms the Global baseline (dashed lines) for almost all cluster sizes, but the gain is most significant at small $k$ (\textit{e.g.}, $k=5$). Higher hardware noise ($p=0.01$, cyan line) suppresses fidelity much faster than lower noise ($p=0.005$, blue line), reducing the viable range of cluster sizes. Therefore, breaking a large network into smaller, independent entangled groups yields a much higher probability of success than attempting global entanglement.

Although the plot confirms that smaller clusters yield higher fidelity, this fragmentation theoretically lowers the security threshold, as fewer cooperating parties are needed to reconstruct a secret compared to a global state. However, this risk is effectively mitigated by the protocol's design:

--Randomized Clustering: The formation of clusters is stochastic, preventing adversaries from predicting which clients will be entangled together in any given round.

--Client Anonymity: Participants in a cluster remain "blind" to one another; they possess their own measurement data but have no knowledge of the identities or locations of the other members in their specific group.

--Dynamic Reconfiguration: By rapidly reshuffling cluster assignments for every key generation round, the system prevents any static group from being targeted, ensuring that even if a small group is compromised, the overall network security remains intact.

In Figure~\ref{fig:compfidelity}, the global model (red surface) illustrates a sharp, non-linear collapse in state fidelity as both the depolarizing noise probability ($p$) and the number of qubits ($N$) increase, dropping to almost zero at higher noise and client levels. The CQSA model (green surface) consistently maintains higher fidelity than the global model across the entire parameter space. 

However, the CQSA approach does not eliminate decoherence; as the noise probability $p$ increases, the green surface clearly tends toward lower fidelity. However, we argue that these results are far more actionable than those of the global model. Because the CQSA framework breaks the problem into independent submodules, the resulting errors are localized rather than systemic. 
\smallskip

\noindent\textbf{Discussion.} CQSA has two inherent limitations arising from its clustered design. First, malicious detection is performed at the level of clusters, and thus, the protocol cannot directly identify or isolate malicious participants within a cluster. However, since cluster formation is randomized across communication rounds, a persistent adversary is unlikely to be grouped with the same set of honest clients repeatedly. Second, CQSA reveals cluster-level aggregated updates to the server, which may introduce privacy leakage when clusters are small. However, this leakage is bound to aggregated information rather than individual updates, and the random reassignment of clients to clusters across rounds further prevents consistent exposure of the same subset of clients.

\section{Conclusion}
\noindent CQSA demonstrates that Byzantine-robust secure aggregation can be made practical for federated learning by replacing global entanglement with randomized, dynamically reconfigured clusters. Our evaluation under depolarizing noise shows that clustering substantially improves the probability of success (fidelity) by localizing decoherence, while the anonymity and reshuffling of the protocol mitigate the reduced risk of privacy (due to the small, limited size of the cluster) that would otherwise arise from fragmentation. Although noise still degrades performance as $p$ increases, the modular structure makes errors more amenable to mitigation techniques that are infeasible at a global scale. Future work can focus on comparing computational overload between the proposed CQSA and present classical secure aggregation protocols,  optimizing cluster sizing under heterogeneous hardware, integrating explicit error-mitigation and purification steps, and validating CQSA on real quantum backends and end-to-end FL workloads.


\bibliographystyle{plain}
\bibliography{sample-base}

\end{document}